\documentclass[11pt,a4paper]{article}
\usepackage[hyperref]{emnlp-ijcnlp-2019}
\usepackage{times}
\usepackage{latexsym}

\usepackage{url}
\usepackage{graphicx}
\usepackage{multirow}
\usepackage{amsthm}
\usepackage{amssymb}
\usepackage{amsmath}
\usepackage{amsfonts}
\usepackage{xcolor}
\usepackage{mdwlist}
\usepackage{microtype}
\usepackage{tabularx}
\usepackage{booktabs}
\usepackage{subcaption}

\usepackage{pifont} 

\newcommand{\one}{\ding{172}}
\newcommand{\two}{\ding{173}}
\newcommand{\three}{\ding{174}}
\newcommand{\four}{\ding{175}}
\newcommand{\five}{\ding{176}}
\newcommand{\six}{\ding{177}}
\newcommand{\seven}{\ding{178}}
\newcommand{\eight}{\ding{179}}
\newcommand{\nine}{\ding{180}}
\usepackage{url}

\aclfinalcopy 

\makeatletter
\renewcommand\outauthor{%
  \begin{tabular}[t]{c}
	\bf\@author
  \end{tabular}}%
\makeatother

\title{FAMULUS: Interactive Annotation and Feedback Generation for Teaching Diagnostic Reasoning}

\author{\bf Jonas Pfeiffer$^1$, Christian M.\ Meyer$^1$, Claudia Schulz$^1$, \\
\textbf{Jan Kiesewetter$^2$, Jan Zottmann$^2$, Michael Sailer$^3$,  Elisabeth Bauer$^3$,} \\ 
\textbf{Frank Fischer$^3$, Martin R. Fischer$^2$, Iryna Gurevych$^1$ }\\
$^1$ Ubiquitous Knowledge Processing (UKP) Lab, Technische Universit\"at Darmstadt, Germany\\
$^2$ Institute of Medical Education, University Hospital, LMU M\"unchen, Germany\\
$^3$ Chair of Education and Educational Psychology, LMU M\"unchen, Germany \\
\texttt {\href{http://famulus-project.de}{http://famulus-project.de}} \\
}

\date{}

\begin{document}
\maketitle
\begin{abstract}   
   Our proposed system FAMULUS helps students learn to diagnose based on automatic feedback in virtual patient simulations, and it supports instructors in labeling training data.
   Diagnosing is an exceptionally  difficult skill to obtain but vital for many different professions (e.g., medical doctors, teachers).
   Previous case simulation systems are limited to multiple-choice questions and thus cannot give constructive individualized feedback on a student's diagnostic reasoning process. 
   Given initially only limited data, we leverage a (replaceable) NLP model to both support experts in their further data annotation with automatic suggestions, and we provide automatic feedback for students.
   We argue that because the central model consistently improves, our interactive approach encourages both students and instructors to recurrently use the tool, and thus accelerate the speed of data creation \emph{and} annotation.
   We show results from two user studies on diagnostic reasoning in medicine and teacher education and outline how our system can be extended to further use cases.

\end{abstract}

\section{Introduction}

\paragraph{Motivation.}
Supporting students in learning has been the life purpose of many teachers throughout history.
With the growing number of people who choose an academic path, it becomes increasingly important to leverage automatic methods to guide students and give them individualized feedback.

However, existing systems for technology-enhanced learning, mostly address skills on recalling, explaining, and applying knowledge, e.g., in automatically generated language learning exercises \citep{Madnani2016} and math word problems \citep{KoncelKedziorski2016}.
More complex cognitive tasks such as diagnostic reasoning require analytic and decision-making skills, for which there are yet only few solutions, even though diagnostic skills are vital for many professions (e.g., medical doctors searching for a therapy, teachers identifying potential mental disorders at an early stage, engineers diagnosing a machine failure, etc.).

Training diagnostic skills is hard and typically relies on time-consuming and hard-to-control live role-plays.
Online case simulations involving so-called \emph{virtual patients} crystallized as an effective alternative to role-playing games \citep{Berman2016,Jin2018}.
In case simulations, students collect information on a virtual patient across multiple screens, e.g., from patient--doctor dialogs, lab results, and medical imaging.
To date, the students formulate their 
final diagnosis by means of multiple-choice questions, which are easy to assess, but prevent important analyses of the effectiveness and the efficiency of the diagnostic reasoning process.
This is why we propose to complement multiple-choice questions with prompts asking for explanations of the students' thought process.
The open-form textual explanations enable good insight into the diagnostic reasoning process rather than only its result, leaving room for constructive methodological feedback. However, the text analysis and feedback generation components are highly complex and require advanced Natural Language Processing (NLP) techniques.

\begin{figure*}
  \centering
  \includegraphics[width=1.0\linewidth]{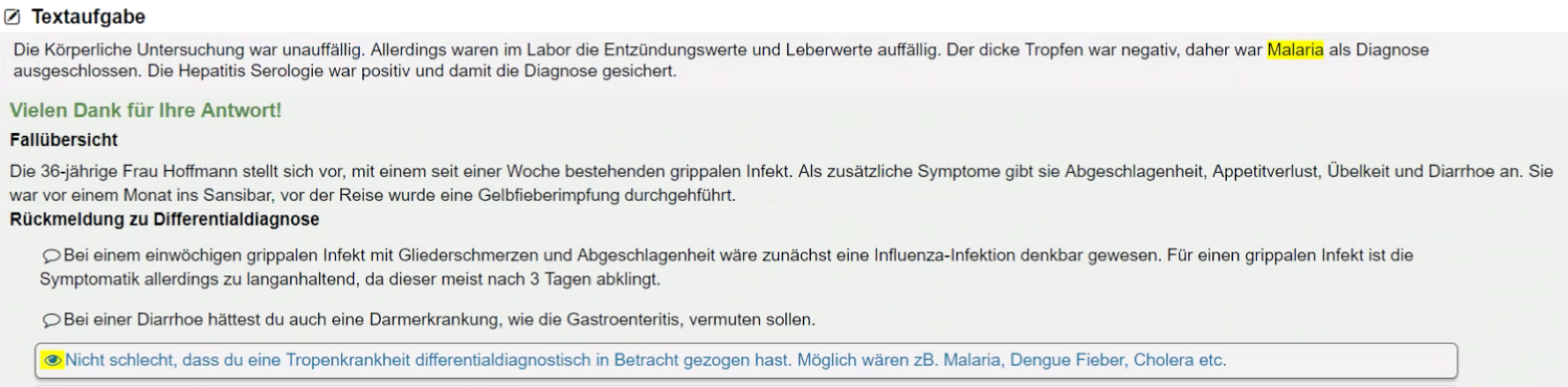}
  \caption{Excerpt of the generated feedback (bottom) to a student's explanation of her/his diagnostic process (top). 
  Blue responses are covered by the text and highlighted in yellow. Black responses are not covered by the diagnosis.}
  \label{fig:feedback}
\end{figure*}

\paragraph{Contributions.}
To tackle this task, we propose our FAMULUS system to generate individual feedback on the students' diagnostic skills.
FAMULUS integrates (a) state-of-the-art neural sequence labeling models to generate individualized feedback, incorporated in our novel NeuralWeb service,
and (b) a corpus construction tool enabling interactive model training with (c) an existing tool for conducting case simulations. 
The backbone of our system is a sequence labeling approach to identify fine-grained diagnostic entities (e.g., \textit{liver values}, \textit{blood pressure}) and epistemic activities \citep{Fischer2014} such as \textit{drawing conclusions} and \textit{evidence evaluation},  in German text.
We have previously published the scientific foundation of our system \citep{Schulz2018,Schulz2018arxiv,Schulz2019aaai}, but introduce the technical aspects of the system for the first time. 
Additionally, for the first time, we evaluate its applicability to real-time use cases.
Our evaluation results on the prediction quality and the annotation effectiveness are based on two user studies with German medicine (Med) and teacher education (TEd) students. 
We show how instructors are relieved from the burden of pre-annotating huge amounts of data by our interactive annotation workflow, and we discuss the generated individualized feedback that helps students acquire diagnostic skills.
Finally, we explain how our system can be obtained, re-used, and extended to further use cases.

\section{Case Simulation Example}
Imagine a Med student training her/his skills with our system.
She/he receives information about a virtual patient:
The 36 years old Ms.\ Hoffmann reports about a common cold lasting for about one week.
In an interview, she mentions abnormal fatigue, diminished appetite, nausea, and diarrhea.
Further questioning reveals that she stayed in Sansibar about a month ago.
Prior to her travel, she was vaccinated against yellow fever.

Based on 
such inputs and further lab results, 
the student explains her/his diagnosis (see Figure~\ref{fig:feedback}):
``Physical examination was clinically unremarkable.
But the lab results show noticeable inflammation markers and liver values.
Thick blood film was negative, therefore Malaria was excluded as a diagnosis. 
Hepatitis serology was positive, which assures the diagnosis.''

In order to automatically provide feedback, we define a set of \emph{diagnostic classes} covering fine-grained diagnostic entities related to the case (e.g., names of diseases, medical examinations, therapies) manually defined by domain experts, and epistemic activities \citep{Fischer2014} that characterize the reasoning process.
As epistemic activity classes, we use \emph{hypothesis generation} (HG; the derivation of possible answers to the problem), \emph{evidence generation} (EG; the derivation of evidence, e.g., through deductive reasoning or observing phenomena), \emph{evidence evaluation} (EE; the assessment of whether and to which degree evidence supports an answer to the problem), and \emph{drawing conclusions} (DC; the aggregation and weighing of evidence and knowledge to derive a final answer to the problem) discussed by \citet{Schulz2019aaai}.


FAMULUS analyzes the previously mentioned diagnostic text and returns feedback on 
multiple
important aspects related to the case.
It successfully detects all aspects
verbalized in the text (e.g., the discussion of tropical diseases; marked in blue in Figure~\ref{fig:feedback}).
Aspects that are not addressed in the text are discussed and provide additional input to what the student has missed  (e.g., that the differential diagnosis should consider a potential bowel disease due to the diarrhea).
For the present example, the student correctly diagnoses a Hepatitis variant (correct would be Hepatitis A), which is positively acknowledged in the generated feedback.
In the supplementary video material, we show two original German diagnostic texts and the corresponding feedback generated by our system.


\section{System Architecture}
\label{sec:SystemArchitecture}

FAMULUS consists of three intercommunicating components introduced in this section.

\subsection{NeuralWeb}
\label{sec:NeuralWeb}
NeuralWeb\footnote{\href{https://www.github.com/UKPLab/emnlp2019-NeuralWeb}{github.com/UKPLab/emnlp2019-NeuralWeb}} is a Python-based web service that communicates with all other components and thus resembles the core of our system.
It is responsible for interactive training and prediction of the diagnostic classes and for the generation of individualized feedback.
We divide its functionality in a \emph{model} and a \emph{feedback DB} part, encompassed by a \emph{wrapper class} that can be easily adapted for new machine learning methods and case studies.





\paragraph{Model.}
The wrapper class includes a loading function which leverages the downstream model architecture and copies the respective weights into memory. 
The supported neural architectures are written in Keras\footnote{\href{www.keras.io}{keras.io}}, and PyTorch\footnote{\href{www.pytorch.org}{pytorch.org}} and are therefore easy to adapt.
NeuralWeb currently provides a recent BiLSTM architecture \citep{reimers2017reporting} implemented in Keras and Flair \cite{akbik2018contextual} implemented in PyTorch, which holds the current state of the art on many sequence-labeling tasks. 
A prediction function of the wrapper pre-processes a text (sentence splitting and tokenization using NLTK) and leverages the pre-trained model to predict and return the diagnostic classes.

NeuralWeb additionally enables automatic retraining of the model within the framework which is useful when new data has been generated and annotated,  improving the model automatically. This functionality is currently implemented for the Keras-based model.

\begin{figure}
  \centering
  \includegraphics[width=\linewidth,trim=1.1cm 0cm 5.3cm 1cm,clip]{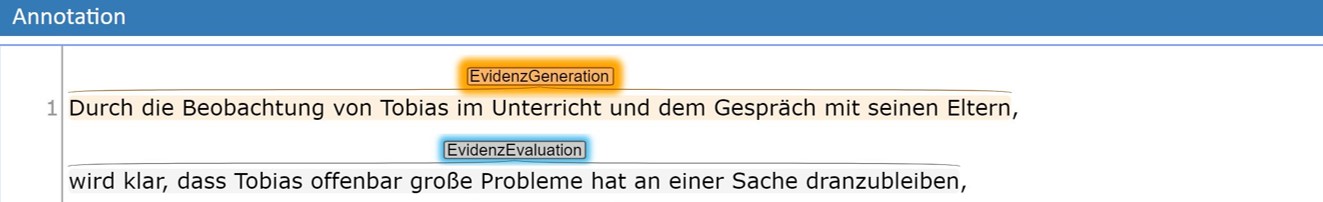}
  \caption{Annotation suggestion (grey) and accepted suggestion (orange) in the INCEpTION platform.}
  \label{fig:inception}
 \end{figure}  
 
\paragraph{Feedback DB.} 
The output of the model is a set of discrete 
diagnostic class labels, which hardly yield valuable feedback for users.
It is thus essential to provide an additional description, indicating whether or not the diagnosis is correct, what is missing, and if the diagnostic process is sound.
We thus introduce a Feedback DB, which includes descriptive text snippets
written by experts.
These descriptions are associated with diagnostic classes 
predicted by the model and a specific case study. 
For example the patient in case 1 \textit{has} Hepatitis A, whereas the patient in case 2 \textit{does not}.
The feedback for a student who diagnoses 
Hepatitis A thus needs to be different with respect to the case she/he currently works on.
The Feedback DB is an independent resource queried by the wrapper class.
With respect to the predicted labels, the corresponding feedback text will be generated.
FAMULUS finally returns the labeled texts spans of the diagnostic text together with the feedback text indicating the reasons for the prediction.

\subsection{INCEpTION}
\label{sec:Inception}
Expert annotation  by instructors is required due to the complexity of labeling diagnostic texts.
For this reason, we leverage the INCEpTION text annotation platform \cite{klie2018} which enables interactive semantic annotation.
The \emph{recommender} system which provides instructors with automatically generated annotation suggestions is one of the key functionalities of the platform. Suggestions can be obtained from various integrated classifiers as well as from external sources such as NeuralWeb. The platform uses the user feedback (accepted/rejected annotations) as well as user-created annotations to continually improve the classifiers. We leverage this functionality to create an efficient interactive annotation process for our diagnostic classes and thus to create training data for our NLP models.
Figure~\ref{fig:inception} shows an example of the labeling process with suggestions by our pre-trained model.

\subsection{CASUS}
\label{sec:Casus}
CASUS\footnote{\href{www.instruct.eu}{www.instruct.eu}} is an interactive system designed for case simulations with virtual patients.
It incorporates all aspects necessary for conducting diagnostic case simulations 
(i.e., videos, images, text, audio integration).
Students  receive information relevant for solving the case.
They are subsequently required to formulate their diagnosis in multiple-choice questions and our new, free-text prompts, directly integrated in CASUS.
After submission, CASUS presents the feedback received from NeuralWeb.

While CASUS is a sophisticated proprietary simulation tool, we would like to stress that this is not a requirement. 
FAMULUS can be used with any open-source front-end tool providing a text box and communicating with NeuralWeb to print out the individualized feedback.
We provide a simple version of such a tool in our GitHub repository together with NeuralWeb.

\section{FAMULUS Process}
\label{sec:Process}

\begin{figure}
  \centering
  \includegraphics[width=1.0\linewidth,trim=0cm .6cm 0cm 0cm,clip]{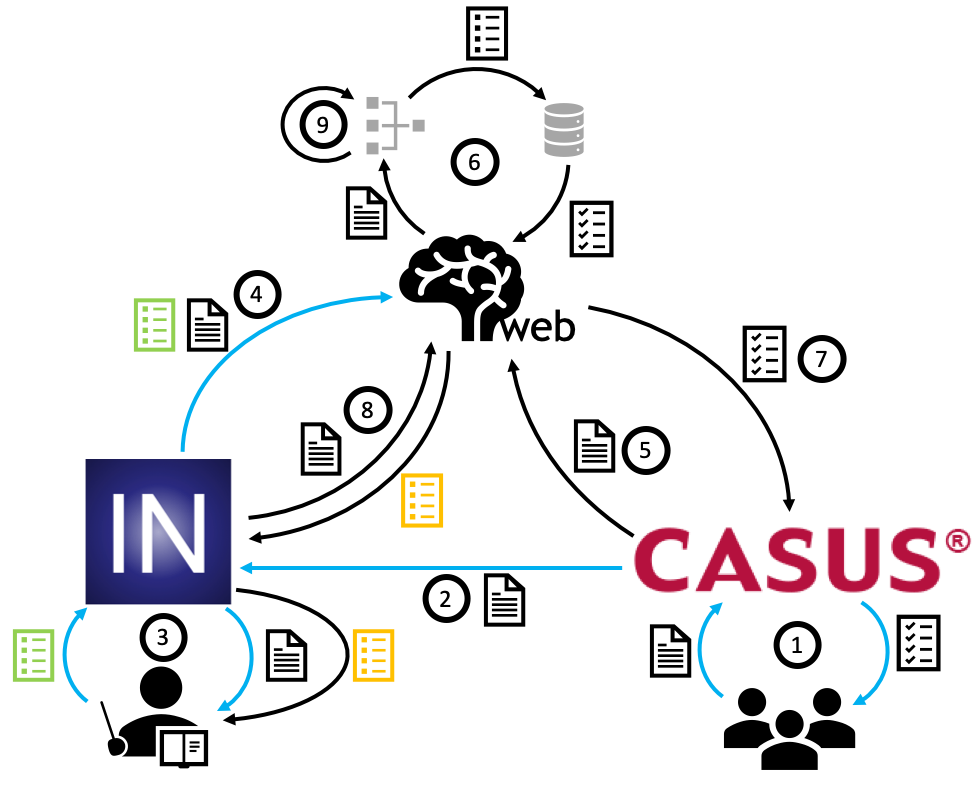}
  \caption{FAMULUS process for annotating data, training models, and generating individualized  feedback. During the cold-start phase, only the blue edges are used, until enough labeled data exists to train a model.}
  \label{fig:process}
\end{figure}

The FAMULUS system consists of an interactive learning cycle connecting the three components introduced in \S \ref{sec:SystemArchitecture} and illustrated in Figure~\ref{fig:process}.

\paragraph{Cold-Start.}
Because a small initial
set of annotated data is necessary to train a preliminary model in NeuralWeb, \one~few pilot 
users first submit their diagnoses to the CASUS system.
In this cold-start phase, the users either do not receive \textit{any} feedback or a \textit{default} feedback text.
For our experiments, all users receive a default gold diagnosis written by experts, for the users to compare their results manually. \two~The students' diagnostic texts are sent to INCEpTION, where \three~ instructors label the data  according to the predefined annotation schema. \four~The gold labels (visualized in green) are stored and sent to NeuralWeb.
Using this labeled training data, we can train our models to automatically predict the diagnostic classes found in a given text. 

\paragraph{Warm-Run.}
After an initial model has been trained, a new set of users can benefit from the trained model to receive individualized feedback. \one~Similarly to the cold-start phase, users work through the case study and submit their diagnosis to the CASUS system.
\five~Instead of receiving a default feedback, the diagnosis is sent to NeuralWeb.
\six~NeuralWeb processes the text through the trained model and generates individualized feedback with regards to the Feedback DB.
\seven~The individualized feedback is sent back to CASUS which visualizes it for the user.

Like in the cold-start phase, \two~the diagnostic text is also sent to INCEpTION.
\eight~But instead of relying solely on the instructor, the trained model in NeuralWeb predicts preliminary annotations (denoted in yellow) which are additionally presented to the instructor (see Figure~\ref{fig:inception}).
\three~These predictions should increase the labeling speed, as in many cases, the instructor simply has to accept the suggestions the model has predicted.
\four~The validated (green) labels are sent back to NeuralWeb and \nine~the model can be interactively retrained 
for each additional data point which has been labeled.

\section{Evaluation}

We employ our proposed FAMULUS architecture in two studies yielding 1,107 Med and 944 TEd diagnostic texts written for eight distinct cases per domain.
While a full analysis of the two studies is beyond the scope of this paper, we focus on three research questions highly relevant to the systemic aspects of FAMULUS:
(1) the quality of the predicted diagnostic classes,
(2) the computation time of the prediction and feedback generation system to assess the applicability of our system in real-time applications, and
(3) the benefits of providing annotation suggestions to the instructors. 



\paragraph{Prediction quality.}

\begin{table}
  \centering
  \begin{tabular}{ll cccc}
    \toprule
    && EG    & EE    & HG    & DC    \\ \midrule
    \multirow{2}{*}{\rotatebox[origin=c]{90}{\textbf{Med}}} & BiLSTM            & 71.60 & 80.20 & 69.28 & 65.32 \\
    & UB & 85.61 & 90.25 & 86.37 & 85.58 \\
    \midrule
    \multirow{2}{*}{\rotatebox[origin=c]{90}{\textbf{TEd}}} & BiLSTM            & 78.53 & 78.87 & 57.16 & 61.77 \\
    & UB & 93.29 & 90.71 & 81.77 & 82.11 \\ \bottomrule
  \end{tabular}
  \caption{Individual macro-F1 scores following \citet{Schulz2019aaai} for each of the epistemic activities. The BiLSTM uses FastText embeddings \cite{FastText}. This architecture is equal to Flair when only using FastText embeddings.
  UB reports the human upper bound (inter-annotator agreement) indicating room for improvement.}
  \label{table:F1-Score}
\end{table}

In Table~\ref{table:F1-Score}, we report the performance of the BiLSTM implementation for predicting epistemic activities in the Med and TEd data.
As we can see, the difficulty of predicting the classes varies between different activities.
Despite some room for improvement with respect to the human upper bound (UB) based on inter-rater agreement, the interactive nature of FAMULUS helps
in succeeding in this attempt by continually improving the model when new data is available. 

We conduct similar experiments for the prediction of fine-grained diagnostic entities, but omit a comprehensive discussion due to space limitations.


\paragraph{Computation time.}

\begin{figure}
  \centering
  \includegraphics[width=1.0\linewidth,trim=0cm .2cm 0cm 0cm,clip]{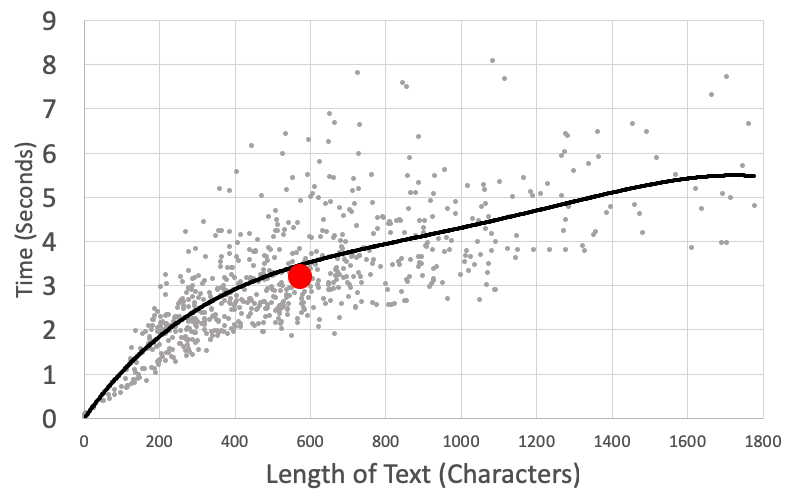}
  \caption{Prediction time for the submitted diagnostic texts of TEd students. The illustrated model is Flair with character and Flair embeddings, hidden size 256, 2 layer BiLSTM with CRF-head. The red dot indicates the mean time and length of 3.15 seconds and 562 characters respectively. The trendline is illustrated in black. The experiment was conducted on a 13-inch 2017 MacBook Pro with i7 Processor and 16GB  RAM.}
  \label{fig:Pred_times}
\end{figure}

In order to present the feasibility of deploying FAMULUS in a real-time scenario,
we plot the inference times of the submitted diagnostic texts in Figure~\ref{fig:Pred_times}.
The inference time includes sentence splitting, 
tokenization, model prediction, and feedback generation using the Feedback DB. We find that on average the submitted texts have a length of 562 characters with an average inference time of 3.15 seconds on a common desktop machine.
The different inference times for similar text lengths are due to variable sentence lengths, as longer sentences require more inference time.
We batch all sentences of one diagnostic text and pass them through the model simultaneously. 
As we can see in the graph, the automatic feedback generation does not surpass 9 seconds.
This is intuitively faster than any human is able to read, process, and output feedback text, even by leveraging prewritten descriptions.
This demonstrates the effectiveness and scalability of FAMULUS in a real-time scenario. 

\paragraph{Annotation suggestions.}
To evaluate the effects of providing annotation suggestions, we have conducted an extensive study \citep{Schulz2019acl} considering annotation time, annotation quality, potential biases, and the ease of use.
%
To this end, we asked five Med and four TEd instructors to annotate diagnostic texts.
Two of the instructors per domain did not receive any suggestions.
For the remaining instructors, we provided suggestions in multiple phases using different models and setups.
Overall, we find positive effects yielding a speed-up of 34 to 42 seconds per annotated text.
The instructors accept 56\,\% (Med) and 54\,\% (TEd) of the annotations.
While we observe a slightly higher inter-annotator agreement if instructors receive suggestions, we also study whether the instructors' decisions are influenced by the suggestions, but only observe a negligible effect.

\section{Dissemination}

In this section, we introduce how the components of our system can be obtained and linked with each other.
Additionally, we discuss multiple use cases that can benefit from our architecture.

\paragraph{Availability.}
The NeuralWeb component is the heart of our system and has been newly developed for our purposes.
We make NeuralWeb available as open-source software in our GitHub repository
under the Apache License 2.0.
We integrate the annotation suggestions generated by our system into the INCEpTION annotation tool, which is available as open-source software under the Apache License 2.0. 
To conduct the case simulations, we use the CASUS system which can be obtained from its publisher Instruct. 
We provide a simple but free alternative to CASUS which includes only the necessary functionality for the FAMULUS system, which is to write the diagnostic text and visualize feedback. This system, together with connection functionalities to INCEpTION and CASUS, can be found in our NeuralWeb repository.

For using FAMULUS, a
server or virtual machine is needed on which the system is deployed.
A thorough description can be found in our GitHub repository, including the respective URLs and ports that need to be adapted.

\paragraph{Use cases.}
Our proposed architecture is primarily useful to prepare and conduct case simulations that train diagnostic skills based on text analysis and automated feedback generation methods.
Besides developing new cases for the Med and TEd domains which is the subject of our research, case simulations can be useful for students in engineering (e.g., diagnosing a machine failure), law (investigating evidence in a lawsuit), economy (optimizing business processes), and many more.

In order to leverage the FAMULUS system, three prior steps need to be made, independent of our system:
(1)~Expert instructors develop a set of case studies, for which they provide all necessary information. The case study can be integrated into a simulation tool such as CASUS or provided in printed form. 
(2)~The instructors define an annotation schema, i.e. what kinds of diagnostic classes should be annotated (e.g., observations of teachers in a classroom).
(3)~As the individualized feedback can vary from case to case, corresponding descriptions need to be defined by the instructors. 


\section{Conclusion}
In this paper, we have introduced FAMULUS, a case simulation system integrating interactive data acquisition and model training, and individualized feedback generation for students' explanations of diagnostic reasoning processes.
Our 
analysis shows how FAMULUS  
helps experts in annotating data fast and reliable while successfully predicting  entities and activities occurring in diagnostic texts.
FAMULUS is applicable in real-time scenarios and generates feedback much faster than humans.
While we focus specifically on diagnostic case simulations in medicine and teacher education, we outline the steps necessary to adapt our approach to many other disciplines requiring the training of diagnostic skills.
%
We open-source all components necessary to employ FAMULUS in new case studies, 
hoping to encourage more research in this area. 


\section*{Acknowledgments}
This work has been supported by the German Federal Ministry of Education and Research (BMBF) under the reference 16DHL1040 (FAMULUS). 

\bibliography{emnlp-ijcnlp-2019}
\bibliographystyle{acl_natbib}

\appendix

\end{document}